\documentclass[11pt]{article}
\usepackage{coling2020,multirow,rotating}
\usepackage{times}
\usepackage{url}
\usepackage{latexsym}
\usepackage{xcolor}
\usepackage{graphicx}

\colingfinalcopy % Uncomment this line for the final submission

\title{AbuseAnalyzer: Abuse Detection, Severity and Target Prediction for Gab Posts}

\author{Mohit Chandra$^{1}$,
Ashwin Pathak$^{2}$, Eesha Dutta$^{3}$\thanks{\ The two authors contributed equally.}\ , 
Paryul Jain$^{4*}$, Manish Gupta$^{5}$ \\ 
\textbf{Manish Shrivastava}$^{6}$, \textbf{Ponnurangam Kumaraguru}$^7$ \\
International Institute of Information Technology, Hyderabad $^{1,2,3,4,5,6}$ \\
Indraprastha Institute of Information Technology, Delhi$^7$ \\
Microsoft$^{1,5}$\\
\tt \{mohit.chandra, eesha.dutta, paryul.jain\}@research.iiit.ac.in \\
\tt ashwin.pathak@alumni.iiit.ac.in, manish.gupta@iiit.ac.in \\ \tt m.shrivastava@iiit.ac.in, pk@iiitd.ac.in}

\date{}

\begin{document}
\maketitle
\begin{abstract}
While extensive popularity of online social media platforms has made information dissemination faster, it has also resulted in widespread online abuse of different types like hate speech, offensive language, sexist and racist opinions, etc. Detection and curtailment of such abusive content is critical for avoiding its psychological impact on victim communities, and thereby preventing hate crimes. 
%technical problem
%related work drawbacks
Previous works have focused on classifying user posts into various forms of abusive behavior. But there has hardly been any focus on estimating the severity of abuse and the target.
%challenges
In this paper, we present a first of the kind dataset with 7601 posts from Gab\footnote{\url{https://Gab.com/}} which looks at online abuse from the perspective of presence of abuse, severity and target of abusive behavior. We also propose a system to address these tasks, obtaining an accuracy of $\sim$80\% for abuse presence, $\sim$82\% for abuse target prediction, and $\sim$65\% for abuse severity prediction.

\iffalse
%Although %a less-moderated platform like 
Gab has a relatively much higher prevalence of abusive content. Surprisingly, the posts have never been previously studied rigorously by the hate speech community.
To demonstrate the efficacy of our proposed methods
The proposed system leads to an accuracy of $\sim$88\% for abuse presence, $\sim$92\% for abuse target prediction, and $\sim$88\% for abuse severity prediction, making the system practically usable.
\fi

\end{abstract}

\section{Introduction}
\label{intro}

In recent times, Online Social Media (OSM) has become an indispensable part of our lives. Not only these websites connect billions of people around the world, but they also serve as a platform for expressing opinions and sharing information quickly. However, recently OSM platforms have been a subject for criticism over the propagation of fake~\cite{shu2017fake} and hateful content~\cite{fortuna2018survey}. Such  cases of online abuse have also translated into real world hate crimes.\footnote{\url{https://www.justice.gov/hatecrimes/hate-crimes-case-examples}}

\iffalse
Fresh,  relevant and humongous content on social media has led to impactful applications like early earthquake warning systems~\cite{sakaki2010earthquake}, epidemic detection systems~\cite{aramaki2011twitter}, pharmacovigilance~\cite{sarker2015utilizing}, etc. 
\fi

%Richness of Abuse in social networks
Abuse in social media is spread across a wide spectrum from mild expressions of attitudes and beliefs to strong violent threats. Inspired by hate theories from Anti-Defamation League (ADL)\footnote{\url{https://www.adl.org/}}, we broadly classify forms of abuse as `Biased Attitude, `Act of Bias and Discrimination' and `Violence and Genocide'. Moreover, abusive content could be targeted at specific individuals (e.g., a politician, a celebrity, etc.) or particular groups (a country, LGBTQ+, a religion, gender, an organization, etc.). 
%``Biased Attitudes'' include trolling, sarcasm, irony, minor accusations. ``Act of Bias and Discrimination'' category includes sexism, racism, xenophobia, homophobia, devaluation and  dehumanizing speech. Finally, ``Violence and Genocide'' category includes violent threats, intimidation, murder and other kinds of extremism.
Detection of such abusive content is critical for avoiding its psychological impact on  victim  communities, and thereby preventing  hate crimes. Prioritization of particular abuse cases can be done if severity of abuse can be automatically assessed. Further, identifying if the abuse target is a person or a large group is critical to predict potential impact set and thereby predict if it could lead to real world crimes along with its scale. Hence, in this paper, we propose three abuse prediction tasks: prediction of abuse presence, abuse severity prediction and abuse target prediction.

%Especially, the increase in cases of online abuse has contributed towards real world hate-crimes \cite{10.1093/bjc/azz049}.

%Prev work -- feature, deep learning, presence/absence, dealt with popular networks like Twitter and Facebook. Gab.
\iffalse
Previous work~\cite{DBLP:conf/naacl/WaseemH16,DBLP:conf/icwsm/DavidsonWMW17} on hate speech detection has leveraged both feature engineering as well a deep learning methods~\cite{DBLP:conf/www/BadjatiyaG0V17}. These studies used Twitter datasets.
\fi

Since traditional OSM websites are reasonably moderated, finding broadly abusive content is possible. But finding abusive behaviour of differing severity is a `needle in a haystack' kind of challenge. In contrast to the other OSM, Gab is relatively unexplored and presents a wider spectrum of online abusive behaviour due to its liberal moderation policy~\cite{DBLP:conf/www/ZannettouBCKSSB18}. Hence, we gathered a dataset from Gab and contribute the labeled posts to the community in the hope of promoting deeper research on abusive content analysis. Gab is an alt-right social media website launched in 2016, which has seen a significant rise in the number of registered users to $1$,$000$,$000$ users along with a daily web traffic of $5.1$ million visits per day by the end of July 2019.\footnote{\url{https://www.similarweb.com/website/gab.com}}

\iffalse
Besides contributing this new dataset, we also propose two other related tasks: abuse severity classification and target detection. Overall, we refer to the three tasks together as abuse detection tasks. Accordingly the contributed Gab dataset has posts labeled for all three aspects: abuse detection, severity and target.

%Gab
 \textcolor{red}{It came in limelight when its name came up in the Pittsburgh synagogue shooting in which the prime suspect had posted anti-Semitic comments just before the incident\footnote{\url{https://edition.cnn.com/2018/10/27/us/pittsburgh-synagogue-active-shooter/index.html}}. It again played a pivotal role in Brazil presidential elections of 2018 when many right-wing Brazilian political groups began promoting their pages on Gab after getting banned on Facebook\footnote{https://www.bloomberg.com/news/articles/2018-10-04/u-s-alt-right-website-lures-brazil-bolsonaro-s-supporters}}. Since then, Gab has attracted a lot of users who have been banned from different OSM networks for violating their hateful content policy~\cite{DBLP:conf/www/ZannettouBCKSSB18}.
 \fi

Our key contributions in this paper are as follows:
\begin{itemize}
    \item We contribute an abuse analysis dataset comprising 7601 Gab posts with finer classification labels associated with presence, severity and target of abuse. The code and dataset are publicly available here\footnote{\url{https://github.com/mohit3011/AbuseAnalyzer}}.
    \item We experiment with traditional machine learning (ML) classifiers with TF-IDF features, for the three abuse prediction tasks. We also experiment with two deep learning (DL) based methods. Our best method leads to high accuracy values of $\sim$80\% for abuse presence, $\sim$82\% for abuse target prediction, and $\sim$65\% for abuse severity prediction.
\end{itemize}

\textbf{Disclaimer}: This paper contains examples of hate content used only for illustrative purposes, reader discretion is advised.

\section{Related Work}

Several past works have explored different kinds of online abuse (like racism, sexism etc.) on traditionally studied platforms like Twitter~\cite{Kwok2013LocateTH,DBLP:conf/naacl/WaseemH16,DBLP:conf/icwsm/DavidsonWMW17,elsherief2018peer} and on some newer web communities like 4chan and Whisper~\cite{DBLP:conf/icwsm/HineOCKLSSB17,DBLP:conf/icwsm/SilvaMCBW16}. But web communities differ from each other through subtleties in language and demographic differences. Gab poses an altogether different challenge as it differs from older web groups primarily in its use of online communities to congregate, organize, and disseminate information in weaponized form \cite{marwick2017media}. Some previous papers~\cite{DBLP:conf/www/ZannettouBCKSSB18,Lima2018InsideTR,DBLP:conf/websci/MathewDG019,DBLP:journals/corr/abs-1809-01644} have presented basic statistical analysis of data extracted from Gab. Recently, ~\newcite{qian-etal-2019-benchmark} presented a dataset of 33,776 posts on Gab annotated on binary labels hate/non-hate. While some papers have focused on racism versus sexism~\cite{DBLP:conf/www/BadjatiyaG0V17}, others have focused on sarcasm, cyber-bullying etc.~\cite{DBLP:conf/websci/FountaCKBVL19}. Initial works in this area focused on feature engineering based methods. With the emergence of deep learning, most of the recent works~\cite{DBLP:conf/websci/FountaCKBVL19,DBLP:conf/acl-alw/SerraLSSBV17,DBLP:conf/acl-alw/ParkF17} have relied on deep learning techniques for abuse detection. To the best of our knowledge, there is no publicly available corpus or prediction system which focuses on fine-grained abusive language classification across all three tasks: prediction of abuse presence, abuse severity prediction and abuse target prediction.

\section{Abuse Severity and Targets}

\subsection{Abuse Severity}
Anti-Defamation League\footnote{\url{https://www.adl.org/sites/default/files/documents/pyramid-of-hate.pdf}} present a pyramid of hate from a sociology perspective (also supported by the ~\newcite{georgescu2018t}).  Although the behaviors at each level of the hate pyramid negatively impact individuals and groups, as one moves up the pyramid, the behaviors have more life-threatening consequences.  Given the five levels in the pyramid, initially we considered five classes of abuse severity. However, we found that some of the classes were infrequent. Moreover, some classes were very similar. Hence, we finally recognized three clear classes of abuse as follows.

\noindent\textbf{Biased Attitude}: Contains posts related to trolling, accusations, sarcasm and insensitive remarks. These posts project minor hate and negation of intimacy towards individual/group through jokes or blaming the target for some mishappening. E.g., ``At lease my choice of president waits till they are out of diapers.'', ``Good news. No wonder they tried to f*ck her nomination over hard.''

\noindent\textbf{Act of Bias and Discrimination}: Consists of posts lying in the category of sexism, racism, xenophobia, homophobia and other forms of abuse which are based on some presumption or stereotype without well founded arguments. Additionally we consider aggression based on the ideologies (leftist, rightist, feminists etc.). This class also cover posts which induce a sense of inferiority complex on the basis of target's physical (`ugly', 'deformed' etc.), mental (`stupid', `cretin' etc.) or moral deficiency (`sl*t', `fraud' etc.). E.g., ``Jews are tragic for the world.'', ``you're are a real f*cking braindead piece of trash b*tchboy.''

\noindent\textbf{Violence and Genocide}: Contains abusive behaviours like violent threat, intimidation and extremism. These posts have statements of intent to inflict harm that would result in serious and lasting bodily impact, where an individual could die or be significantly injured. It reflects intense anger/fear towards the targeted individual/group and depicts them as a threat to society. In many cases, the posts refer to adverse activities like deportations etc. E.g., ``antiwhite whites deserve public hangings'', ``I know what is going on!  I f*ck his mammy and I hurt her. Sorry mam!''

\subsection{Abuse Targets}

A comment targeted towards a particular community impacts a larger audience as compared to one targeted towards a particular individual. Hence, it is important to predict target of the abusive post as one of these three classes.

\noindent\textbf{Individual (Second-Person)}: Targets the person being mentioned in the post. Generally, there is a usage of terms like `@username', `you' and `your' to refer the target. E.g., ``No, but I do realize that you're full of sh*t and know it.'', ``@username is serving a purpose or just a load of hot air.''

\noindent\textbf{Individual (Third-Person)}: Target a third person. Usually, these posts use terms like `he', `she', etc. or many a times the posts mention the name/username of the target. E.g., ``His predatory sexual behavior is still evident.'', ``Another pedophile circles the wagons.''

\noindent\textbf{Group}: Target a group/organization based on ideologies, race, gender, religion, work industry or some other basis. Such posts contain terms like `you all', 'they' or many a times refers to the group in an indirect manner. E.g., ``We have some shit stirrers afoot today. Ignore them'', ``Why not set dead muslims on the curb in a trash bag?''

\section{AbuseAnalyzer Dataset and Results}
\label{sec:dataset}
Our dataset contains 7601 Gab posts classified on three different aspects: abuse presence or not, abuse severity and abuse target. Of the 4120 abusive posts, distribution based on severity is -- `Biased Attitude': 1830, `Act of Bias and Discrimination': 1807, and `Violence and Genocide': 483. For the target classes -- 389 are in `Individual (Second-Person)', 1330 in `Individual (Third-Person)', and 2401 in the `Group' class. The code and dataset are publicly available here\footnote{\url{https://github.com/mohit3011/AbuseAnalyzer}}.

\noindent\textbf{Data Extraction and Pre-processing}: We obtained a collection of 8.4 million Gab posts from \url{http://files.pushshift.io/gab/} for a period of 4 months from Jul to Oct 2018. We used a high precision lexicon  %(available here\footnote{\url{https://www.dropbox.com/sh/i2j0fq35gcjp99h/AAB1eyqUEbbmOt4yqwpy7Njna?dl=0&preview=keyword_list.txt}}) 
which consists of racial, sexist, xenophobic, extremist and other derogatory terminologies aggregated from multiple source.\iffalse
The lexicon has been gathered by aggregating from multiple sources\footnote{\url{https://en.wikipedia.org/wiki/List_of_ethnic_slurs}}$^,$\footnote{\url{https://github.com/t-davidson/hate-speech-and-offensive-language/blob/master/lexicons/refined_ngram_dict.csv}}$^,$\footnote{\url{https://hatebase.org/search_results/language_id\%3Deng}}
\fi We used this to filter 7601 posts written in English for the annotation process. While we made efforts to strike a balance between abusive versus non-abusive posts, we made no efforts to maintain balance within abuse severity or abuse target classes.

\noindent\textbf{Annotation Procedure}: Four annotators with fluent English skills were provided clear guidelines (refined iteratively) for annotating the posts across all the three abuse prediction tasks. In case a post could belong to more than one severity classes, annotators were asked to mark the higher severity class (based on life-threatening consequences), to avoid multi-labels. Each example was annotated by exactly 3 annotators and all the disagreements were resolved after involving all the annotators. As a measure of inter-annotator agreement, we observed Cohen's Kappa Score~\cite{cohen1960} as (1) 0.719 for presence/absence of abuse, (2) 0.720 for presence+target, and (3) 0.683 for presence+severity classification. In each case the Kappa score is near 0.7 which is a very good agreement among the annotators.

\noindent\textbf{Dataset Statistics and Analysis}: Table~\ref{tab:stats} shows the distribution of the `Target' labels among each of the `Severity' classes. We observe that majority of the abusive posts are against the `Group' class, specifically for `Act of Bias and Discrimination' class which is intuitive since this category covers the topics of racism, sexism etc. 

\begin{table}[!ht]%
\centering
\small
\begin{tabular}{|l|l|l|l|l|}
\hline
Severity\ $\downarrow$\ \ \ \ \ \ \ \ \ \ \  Target\ $\rightarrow$ &Individual Second P.&Individual Third P.&Group&Total\\
\hline
Biased Attitude &226&650&954&1830\\
\hline
Act of Bias and Discrimination&129&543&1135&1807\\
\hline
Violence and Genocide&34&137&312&483\\
\hline
Total&389&1330&2401&4120\\
\hline
\end{tabular}
\caption{Distribution of posts across various abuse severity and abuse target classes.}
\label{tab:stats}
\end{table}

Table~\ref{tab:ngrams} shows popular unigrams and bigrams for  various severity and target classes. We observe that: (1) Community related words and bigrams like `jew',  `muslim', etc. are quite frequent for `Act of Bias and Discrimination' class which is in line with the nature of posts on Gab. (2) violent ngrams like `kill', `the holocaust' are present in the `Violence and Genocide' class. (3) Second person pronouns like ``you'', ``yourself'', etc. are frequent in the `Individual (Second-Person)' class. (4) Third person pronouns and bigrams like ``he'', ``she'', ``hes a'', etc. are frequent in the `Individual (Third-Person)' class. (5) Multiplicity indicating ngrams like ``these people'', ``them'', etc. are popular in the `Group' class.

\begin{table*}[!ht]%
\centering
\small
\begin{tabular}{|l|p{0.13\textwidth}|p{0.33\textwidth}|p{0.4\textwidth}|}
\hline
\multicolumn{2}{|c|}{}&Unigrams&Bigrams\\
\hline
\hline
\multirow{3}{*}{\rotatebox{90}{Severity\ \ \ \ \ }}&Biased Attitude & lol, white, f*ck, against, killed, twitter, government, usermention, america & you are, they are, trying to, illegal alien, going to, to do, to get \\
\cline{2-4}
&Act of Bias and Discrimination&jews, white, black, muslims, stupid, islam, b*tch, k*ke, evil, rape & you are, the jews, of sh*t, jews are, white people, muslims are, a race, white people, a n*gger, a k*ke\\
\cline{2-4}
&Violence and Genocide&f*ck, kill, hell, die, b*tch, lol, fight, muslims, white, war&to hell, the f*ck, to kill, the b*tch, rid of, kill all, get rid, f*ck the, to die, the holocaust\\
\hline
\multirow{3}{*}{\rotatebox{90}{Target\ \ \ \ \ \ \ \ }}&Second person&you, your, youre, f*ck, stupid, sh*t, jew, b*tch, yourself, @username & you are, if you, are you, you don’t, do you, youre a, you just, your own\\
\cline{2-4}
&Third person&he, her, she, his, you, this, b*tch, sh*t, trump, him, @username&she is, he is, hes a, he was, a jew, she was, he has, illegal alien\\
\cline{2-4}
&Group&they, you, all, their, jews, them, people, f*ck, white, sh*t&they are, the jews, the left, jews are, these people, white people, they will, the US, all of, all the\\
\hline
\end{tabular}
\caption{Frequent unigrams and bigrams for each of the abuse severity and abuse target classes.}
\label{tab:ngrams}
\end{table*}

\begin{figure}[h]
  \centering
  \includegraphics[width=0.7\columnwidth]{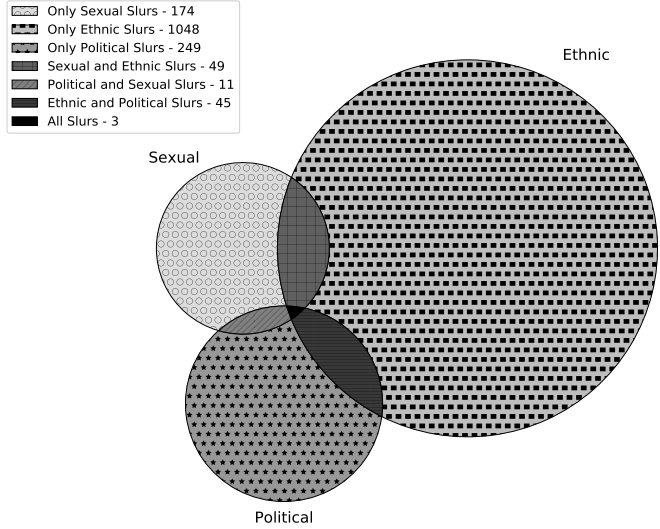}
  \caption{Frequency of slur words in Gab posts. The legend indicates \#posts. \iffalse(best viewed in color)\fi}
  \label{fig:vennDiagram}
\end{figure}

As a final dataset analysis step, we wished to look at frequency of sexual, political and ethnic slurs across various abusive posts in our dataset. Figure~\ref{fig:vennDiagram} shows the frequency in a Venn diagram. Our dictionaries have the following sizes: sexual (51), political (13), and ethnic (131)\footnote{\url{https://en.wikipedia.org/wiki/List_of_ethnic_slurs}} (manually curated from multiple sources). There are also instances of posts having abusive slurs from more than one classes as shown in the figure. We observe that most posts contain ethnic slurs followed by political slurs. This again validates the alt-right nature of the platform. Among political slurs, we observe most of them were against democrats.

\section{Experiments}

\noindent\textbf{Prediction Results}: We experiment with multiple statistical ML methods (Support Vector Machines (SVM), XGBoost and Logistic Regression (LR)) using TF-IDF features. We also trained two Deep Learning based models: (1) Bidirectional Encoder Representations from Transformers (BERT)~\cite{devlin2018bert} using transfer learning and (2)  GloVe-based~\cite{DBLP:conf/emnlp/PenningtonSM14} Long Short Term Memory~\cite{Hochreiter:1997:LSM:1246443.1246450} networks (referred as GloVe+LSTM). With BERT, we use an additional 2-layer multi-layer Perceptron (MLP) for classification with a dropout value of 0.2.% (code available here \footnote{\url{https://www.dropbox.com/sh/9bszxfidv7i1ju8/AABMV-4yEf1klI78fYikeQGSa?dl=0}}). 
We trained both the DL networks using Adam optimizer~\cite{kingma2014adam}. Table~\ref{tab:mldlresults} shows 5-fold cross validation accuracy (micro F1) and macro F1 for each of the methods. We observe that our BERT based model outperforms other methods with SVM being the best out of the ML models.  

\setlength{\tabcolsep}{2pt}
\begin{table}[h]
    \centering
    \small
    \begin{tabular}{|l|c|c|c|c|c|c|}
    \hline
&\multicolumn{2}{c|}{Presence}&\multicolumn{2}{c|}{Target prediction}&\multicolumn{2}{c|}{Severity prediction}\\
\hline
Classifier&Macro F1&Micro F1/Acc&Macro F1&Micro F1/Acc&Macro F1&Micro F1/Acc\\
\hline
\hline
SVM & $0.7277\pm 0.0112$ & $0.7279\pm 0.0113$ &$0.7085\pm 0.0207$ & $0.7619\pm 0.0120$&$0.5787\pm 0.0211$ & $0.6238\pm 0.0236$\\
\hline
XGBoost & $0.7157\pm 0.0097$ & $0.7165\pm 0.0096$ &$0.6750\pm 0.0236$ & $0.7405\pm 0.0126$&$0.5296\pm 0.0141$ & $0.6238\pm 0.0084$\\
\hline
LR & $0.7235\pm 0.0135$ & $0.7239\pm 0.0135$ &$0.6961\pm 0.0185$ & $0.7558\pm 0.0094$&$0.5674\pm 0.0132$ & $0.6201\pm 0.0168$\\
\hline
BERT& $0.7985\pm 0.0110$ & $0.8015\pm 0.0105$ &$0.7893\pm 0.0104$ & $0.8201\pm 0.0086$&$0.6244 \pm 0.0465$ & $0.6502\pm 0.0509$\\
\hline
GloVe+LSTM & $0.5261\pm 0.2365$ & $0.6396\pm 0.1332$ &$0.4009\pm 0.0324$ & $0.6097\pm 0.0097$&$0.4253\pm 0.0480$ & $0.4726\pm 0.0150$\\
\hline
    \end{tabular}
    \caption{AbuseAnalyzer Results for Presence, Target and Severity prediction across multiple classifiers.}
    \label{tab:mldlresults}
\end{table}

\begin{table}[!ht]%
\small
\begin{minipage}[b]{0.43\linewidth}
\centering
\begin{tabular}{|l|l|l|l|l|}
\hline
\multicolumn{2}{|c|}{}&\multicolumn{3}{c|}{Predicted}\\
\cline{3-5}
\multicolumn{2}{|c|}{}&Second-Person&Third-Person& Group\\
\hline
\hline
\multirow{3}{*}{\rotatebox{90}{Actual\ \ }}&Second-Person&319&34&36\\
\cline{2-5}
&Third-Person&61&1078&191\\
\cline{2-5}
&Group&111&308&1982\\
\hline
\end{tabular}
\caption{Confusion matrix for Abuse Target prediction using BERT.}
\label{tab:cm_target}
\end{minipage}
\hspace{0.02\linewidth}
\begin{minipage}[b]{0.55\linewidth}
\centering
\begin{tabular}{|l|l|l|l|l|}
\hline
\multicolumn{2}{|c|}{}&\multicolumn{3}{c|}{Predicted}\\
\cline{3-5}
\multicolumn{2}{|c|}{}&Biased Attitude&\vtop{\hbox{\strut Act of Bias and}\hbox{\strut Discrimination}}  &\vtop{\hbox{\strut Violence and}\hbox{\strut  Genocide}}\\
\hline
\hline
\multirow{3}{*}{\rotatebox{90}{Actual\ \ \ \ }}&Biased Attitude&1252&386&192\\
\cline{2-5}
&\vtop{\hbox{\strut Act of Bias and}\hbox{\strut Discrimination}}&503&1104&200\\
\cline{2-5}
&\vtop{\hbox{\strut Violence and}\hbox{\strut  Genocide}}&98&63&322\\
\hline
\end{tabular}
\caption{Confusion matrix for Abuse Severity prediction using BERT.}
\label{tab:cm_severity}
\end{minipage}
\end{table}

\vspace{4pt}

\noindent\textbf{Confusion matrices}: We show confusion matrix for abuse target and severity prediction tasks in Tables~\ref{tab:cm_target} and~\ref{tab:cm_severity} respectively. The entries denote the sum of examples in the 5-fold cross validation.

Table~\ref{tab:cm_target} shows the confusion matrix for the task of `Abuse Target' prediction. As observed, a good chunk of examples annotated in the `Group' category have been classified in the `Individual (Third-Person)' category due to the presence of many named entities in the `Individual (Third-Person)' category which can easily be confused with a group name. Similar is the case with the instances belonging to the `Individual (Third-Person)' class but classified in the `Group' category. Table~\ref{tab:cm_severity} presents the confusion matrix for the task of Abuse Severity prediction. We observed a major miss-classification in the case when the ground truth was `Act of Bias and Discrimination' but the predicted label was in the category of `Biased Attitude', one of the main reason behind this error can be the closeness in these two categories. Furthermore the subjectivity attached to behaviours like Trolling, Sarcasm (present in `Biased Attitude') and Devaluation, Dehumanization speech (present in `Act of Bias and Discrimination') causes confusion for the classifier, in fact this was even observed during the annotation procedure where the annotator had subjective disagreements on classifying the posts in the aforementioned behaviours. 

\section{Error Analysis}
  
Table~\ref{tab:model_fail} presents the cases where AbuseAnalyzer mis-classifies the examples. We present some interesting cases for each of the three abuse prediction tasks. For the task of prediction of presence of abuse, we see that terms like `black', `muslims' which are prone to online abuse pose a challenge for the classifier. For example, the first post in Table~\ref{tab:model_fail} talks about the adoption of a girl belonging to the black community, this example is non-abusive but it is wrongly classified as abusive due to the presence of potentially racial terms. Similar is the case with the second post which reports a news of arrest of \textit{muslim jihadists}. 

In example 4 in Table~\ref{tab:model_fail} the presence of the pronoun `you' along with the overall tone of the post of being sarcastic confused the system to predict the target class as `Individual (Second-Person)' where the ground truth label `Group' as the post conveys a racist ideology against Jews. Example 5 presents an interesting case which trolls the concerned person while making a general statement about the world, due to the presence of terms like `evil' along with `world', the system gets confused. In example 6 the reference to the third person has been made using `@username' but later the pronoun `you' has been used to refer to this person, this change in the way of referencing confused the system. 

Example 7 in Table~\ref{tab:model_fail} is a sexist comment on the target which blames her for making false accusation of rape. But the presence of an extremist term like \textit{rape} made the classifier to commit error. Example 8 presents a case of an extremist post which propagates the hate in a subtle way. The post talks about killing immigrants from across the border. This phenomenon was common with other posts where the hate was expressed in a very subtle way without using and explicit terms. In example 9 we have a case of trolling, where the person posting has trolled national socialists. 

\begin{table}[h]
    \small
    \centering
\begin{tabular}{|l|p{0.55\columnwidth}|p{0.2\columnwidth}|p{0.18\columnwidth}|}
\hline
Task&Post&Our Prediction&Ground Truth\\
\hline
\hline
\multirow{3}{*}{\rotatebox{90}{Presence\ \ \ \ \ \ \ \ }}&Under a video about a black girl finding out shes being adopted by her white parents&Abusive&Non-Abusive\\
\cline{2-4}
&On Friday the FBI arrested all five Muslim jihadists connected to the New Mexico compound The suspects were charged with violating federal firearms and conspiracy laws.&Abusive&Non-Abusive\\
\cline{2-4}
&Thats how you know she is down for anything&Non-Abusive&Abusive\\
\hline
\multirow{3}{*}{\rotatebox{90}{Target \ \ \ \ \ \ \ \ \ \ \ \ }}&Ill give you a little secret Go after Jews You wouldnt last minutes I didnt&Individual Second Person&Group\\
\cline{2-4}
&You just cant make this crap up Evil sure has a strong presence in this world.&Group&Individual Second Person\\
\cline{2-4}
&My tweet to this creature usermention You scrubbed your Social Media history but its too late The FBI is investigating you now You better lawyer up You wont do well in Prison.&Individual Second Person&Individual Third Person\\
\hline
\multirow{3}{*}{\rotatebox{90}{Severity\ \ \ \ \ \ \ \ \ }}&Rape Im sure she was begging for it Doesnt look like a rape scene to me&Violence and Genocide&Act of Bias and Discrimination\\
\cline{2-4}
&As immigrants flow across US border American guns go south&Act of Bias and Discrimination&Violence and Genocide\\
\cline{2-4}
&How do yall national socialists feel now that the democrats are adopting national socialist policies instead of marxist policies&Act of Bias and Discrimination&Biased Attitude\\
\hline
\end{tabular}
\caption{Sample cases where AbuseAnalyzer predicts incorrectly in comparison to the ground truth.}
\label{tab:model_fail}
\end{table}

\section{Conclusion}
In this paper, we presented a novel dataset with 7601 Gab posts labeled for abuse presence, target and severity. We experimented with both statistical and deep learning based models for each of these tasks and showed that the BERT based model performs the best. Additionally, we presented detailed analysis on the proposed BERT based method with the help of confusion matrices and error analysis.

There are several open avenues for the presented work like exploring context based abuse detection especially in social media post and reply threads. Another interesting direction can be to use data from multiple modalities like images, videos and speech along with the text for the task of abuse detection.

% include your own bib file like this:
\bibliographystyle{coling}
\bibliography{coling2020}

\end{document}